\documentclass[a4paper,twoside]{article}

\usepackage{epsfig}
\usepackage{subcaption}
\usepackage{calc}
\usepackage{comment}
\usepackage{graphicx}

\usepackage{amsmath}
\usepackage{subcaption}
\usepackage{amssymb}
\usepackage{hyperref}
\usepackage{algpseudocode}
\usepackage{booktabs}
\usepackage{comment}
\usepackage{caption} 
\usepackage{times}
\usepackage{epsfig}
\usepackage{adjustbox}

\usepackage{multirow}
\usepackage{times}
\usepackage{colortbl}
\usepackage{epsfig}
\usepackage{graphicx}
\usepackage{adjustbox}
\usepackage{amsmath}
\usepackage{amssymb}

\usepackage[utf8]{inputenc} 
\usepackage[T1]{fontenc}    
\usepackage{url}            
\usepackage{booktabs}       
\usepackage{amsfonts}       
\usepackage{nicefrac}       
\usepackage{bbm}            
\usepackage{enumitem}
\usepackage{float}
\usepackage{microtype}
\usepackage[ruled,vlined,linesnumbered]{algorithm2e}
\usepackage{enumitem}
\usepackage{amssymb}
\usepackage{amstext}
\usepackage{amsmath}
\usepackage{amsthm}
\usepackage{multicol}
\usepackage{pslatex}
\usepackage{apalike}
\usepackage{algorithm2e}
\usepackage[bottom]{footmisc}
\usepackage{SCITEPRESS}     
\newcommand{\base}{\mathrm{base}}

\newcommand{\train}{\mathrm{train}}
\newcommand{\test}{\mathrm{test}}
\newcommand{\val}{\mathrm{val}}

\begin{document}

\title{Image compositing is all you need for data augmentation}

\author{\authorname{Ang Jia Ning Shermaine\sup{1}, Michalis Lazarou\sup{2} and Tania Stathaki\sup{1}}
\affiliation{\sup{1}Imperial College London, U.K.}
\affiliation{\sup{2}University of Surrey, U.K.}
}

\keywords{data augmentation, image classification, generative modelling, Stable Diffusion, ControlNet}

\newcommand{\head}[1]{{\smallskip\noindent\textbf{#1}}}
\newcommand{\alert}[1]{{\color{red}{#1}}}
\newcommand{\mich}[1]{\textcolor{blue}{#1}}

\newcommand{\sm}{\scriptsize}
\newcommand{\eq}[1]{(\ref{eq:#1})}

\newcommand{\Th}[1]{\textsc{#1}}
\newcommand{\mr}[2]{\multirow{#1}{*}{#2}}
\newcommand{\mc}[2]{\multicolumn{#1}{c}{#2}}
\newcommand{\tb}[1]{\textbf{#1}}
\newcommand{\ch}{\checkmark}

\newcommand{\red}[1]{{\color{red}{#1}}}
\newcommand{\blue}[1]{{\color{blue}{#1}}}
\newcommand{\green}[1]{{\color{green}{#1}}}
\newcommand{\gray}[1]{{\color{gray}{#1}}}

\newcommand{\citeme}[1]{\red{[XX]}}
\newcommand{\refme}[1]{\red{(XX)}}

\newcommand{\fig}[2][1]{\includegraphics[width=#1\columnwidth]{fig/#2}}
\newcommand{\figh}[2][1]{\includegraphics[height=#1\columnwidth]{fig/#2}}


\newcommand{\tran}{^\top}
\newcommand{\mtran}{^{-\top}}
\newcommand{\zcol}{\mathbf{0}}
\newcommand{\zrow}{\zcol\tran}

\newcommand{\ind}{\mathbbm{1}}
\newcommand{\expect}{\mathbb{E}}
\newcommand{\nat}{\mathbb{N}}
\newcommand{\zahl}{\mathbb{Z}}
\newcommand{\real}{\mathbb{R}}
\newcommand{\proj}{\mathbb{P}}
\newcommand{\prob}{\mathbf{Pr}}
\newcommand{\normal}{\mathcal{N}}

\newcommand{\mif}{\textrm{if}\ }
\newcommand{\other}{\textrm{otherwise}}
\newcommand{\minimize}{\textrm{minimize}\ }
\newcommand{\maximize}{\textrm{maximize}\ }
\newcommand{\st}{\textrm{subject\ to}\ }

\newcommand{\id}{\operatorname{id}}
\newcommand{\const}{\operatorname{const}}
\newcommand{\sgn}{\operatorname{sgn}}
\newcommand{\var}{\operatorname{Var}}
\newcommand{\mean}{\operatorname{mean}}
\newcommand{\trace}{\operatorname{tr}}
\newcommand{\diag}{\operatorname{diag}}
\newcommand{\vect}{\operatorname{vec}}
\newcommand{\cov}{\operatorname{cov}}
\newcommand{\sign}{\operatorname{sign}}
\newcommand{\prj}{\operatorname{proj}}

\newcommand{\softmax}{\operatorname{softmax}}
\newcommand{\clip}{\operatorname{clip}}

\newcommand{\defn}{\mathrel{:=}}
\newcommand{\peq}{\mathrel{+\!=}}
\newcommand{\meq}{\mathrel{-\!=}}

\newcommand{\floor}[1]{\left\lfloor{#1}\right\rfloor}
\newcommand{\ceil}[1]{\left\lceil{#1}\right\rceil}
\newcommand{\inner}[1]{\left\langle{#1}\right\rangle}
\newcommand{\norm}[1]{\left\|{#1}\right\|}
\newcommand{\abs}[1]{\left|{#1}\right|}
\newcommand{\frob}[1]{\norm{#1}_F}
\newcommand{\card}[1]{\left|{#1}\right|\xspace}
\newcommand{\diff}{\mathrm{d}}
\newcommand{\der}[3][]{\frac{d^{#1}#2}{d#3^{#1}}}
\newcommand{\pder}[3][]{\frac{\partial^{#1}{#2}}{\partial{#3^{#1}}}}
\newcommand{\ipder}[3][]{\partial^{#1}{#2}/\partial{#3^{#1}}}
\newcommand{\dder}[3]{\frac{\partial^2{#1}}{\partial{#2}\partial{#3}}}

\newcommand{\wb}[1]{\overline{#1}}
\newcommand{\wt}[1]{\widetilde{#1}}

\def\xssp{\hspace{1pt}}
\def\ssp{\hspace{3pt}}
\def\msp{\hspace{5pt}}
\def\lsp{\hspace{12pt}}

\newcommand{\cA}{\mathcal{A}}
\newcommand{\cB}{\mathcal{B}}
\newcommand{\cC}{\mathcal{C}}
\newcommand{\cD}{\mathcal{D}}
\newcommand{\cE}{\mathcal{E}}
\newcommand{\cF}{\mathcal{F}}
\newcommand{\cG}{\mathcal{G}}
\newcommand{\cH}{\mathcal{H}}
\newcommand{\cI}{\mathcal{I}}
\newcommand{\cJ}{\mathcal{J}}
\newcommand{\cK}{\mathcal{K}}
\newcommand{\cL}{\mathcal{L}}
\newcommand{\cM}{\mathcal{M}}
\newcommand{\cN}{\mathcal{N}}
\newcommand{\cO}{\mathcal{O}}
\newcommand{\cP}{\mathcal{P}}
\newcommand{\cQ}{\mathcal{Q}}
\newcommand{\cR}{\mathcal{R}}
\newcommand{\cS}{\mathcal{S}}
\newcommand{\cT}{\mathcal{T}}
\newcommand{\cU}{\mathcal{U}}
\newcommand{\cV}{\mathcal{V}}
\newcommand{\cW}{\mathcal{W}}
\newcommand{\cX}{\mathcal{X}}
\newcommand{\cY}{\mathcal{Y}}
\newcommand{\cZ}{\mathcal{Z}}
\newcommand{\cPi}{\mathcal{\pi}}

\newcommand{\vA}{\mathbf{A}}
\newcommand{\vB}{\mathbf{B}}
\newcommand{\vC}{\mathbf{C}}
\newcommand{\vD}{\mathbf{D}}
\newcommand{\vE}{\mathbf{E}}
\newcommand{\vF}{\mathbf{F}}
\newcommand{\vG}{\mathbf{G}}
\newcommand{\vH}{\mathbf{H}}
\newcommand{\vI}{\mathbf{I}}
\newcommand{\vJ}{\mathbf{J}}
\newcommand{\vK}{\mathbf{K}}
\newcommand{\vL}{\mathbf{L}}
\newcommand{\vM}{\mathbf{M}}
\newcommand{\vN}{\mathbf{N}}
\newcommand{\vO}{\mathbf{O}}
\newcommand{\vP}{\mathbf{P}}
\newcommand{\vQ}{\mathbf{Q}}
\newcommand{\vR}{\mathbf{R}}
\newcommand{\vS}{\mathbf{S}}
\newcommand{\vT}{\mathbf{T}}
\newcommand{\vU}{\mathbf{U}}
\newcommand{\vV}{\mathbf{V}}
\newcommand{\vW}{\mathbf{W}}
\newcommand{\vX}{\mathbf{X}}
\newcommand{\vY}{\mathbf{Y}}
\newcommand{\vZ}{\mathbf{Z}}

\newcommand{\va}{\mathbf{a}}
\newcommand{\vb}{\mathbf{b}}
\newcommand{\vc}{\mathbf{c}}
\newcommand{\vd}{\mathbf{d}}
\newcommand{\ve}{\mathbf{e}}
\newcommand{\vf}{\mathbf{f}}
\newcommand{\vg}{\mathbf{g}}
\newcommand{\vh}{\mathbf{h}}
\newcommand{\vi}{\mathbf{i}}
\newcommand{\vj}{\mathbf{j}}
\newcommand{\vk}{\mathbf{k}}
\newcommand{\vl}{\mathbf{l}}
\newcommand{\vm}{\mathbf{m}}
\newcommand{\vn}{\mathbf{n}}
\newcommand{\vo}{\mathbf{o}}
\newcommand{\vp}{\mathbf{p}}
\newcommand{\vq}{\mathbf{q}}
\newcommand{\vr}{\mathbf{r}}
\newcommand{\Vs}{\mathbf{s}}
\newcommand{\vt}{\mathbf{t}}
\newcommand{\vu}{\mathbf{u}}
\newcommand{\vv}{\mathbf{v}}
\newcommand{\uu}{\mathbf{u}}
\newcommand{\cc}{\mathbf{c}}
\newcommand{\vw}{\mathbf{w}}
\newcommand{\vx}{\mathbf{x}}
\newcommand{\vy}{\mathbf{y}}
\newcommand{\vz}{\mathbf{z}}

\newcommand{\vone}{\mathbf{1}}
\newcommand{\vzero}{\mathbf{0}}

\newcommand{\valpha}{{\boldsymbol{\alpha}}}
\newcommand{\vbeta}{{\boldsymbol{\beta}}}
\newcommand{\vgamma}{{\boldsymbol{\gamma}}}
\newcommand{\vdelta}{{\boldsymbol{\delta}}}
\newcommand{\vepsilon}{{\boldsymbol{\epsilon}}}
\newcommand{\vzeta}{{\boldsymbol{\zeta}}}
\newcommand{\veta}{{\boldsymbol{\eta}}}
\newcommand{\vtheta}{{\boldsymbol{\theta}}}
\newcommand{\viota}{{\boldsymbol{\iota}}}
\newcommand{\vkappa}{{\boldsymbol{\kappa}}}
\newcommand{\vlambda}{{\boldsymbol{\lambda}}}
\newcommand{\vmu}{{\boldsymbol{\mu}}}
\newcommand{\vnu}{{\boldsymbol{\nu}}}
\newcommand{\vxi}{{\boldsymbol{\xi}}}
\newcommand{\vomikron}{{\boldsymbol{\omikron}}}
\newcommand{\vpi}{{\boldsymbol{\pi}}}
\newcommand{\vrho}{{\boldsymbol{\rho}}}
\newcommand{\vsigma}{{\boldsymbol{\sigma}}}
\newcommand{\vtau}{{\boldsymbol{\tau}}}
\newcommand{\vupsilon}{{\boldsymbol{\upsilon}}}
\newcommand{\vphi}{{\boldsymbol{\phi}}}
\newcommand{\vchi}{{\boldsymbol{\chi}}}
\newcommand{\vpsi}{{\boldsymbol{\psi}}}
\newcommand{\vomega}{{\boldsymbol{\omega}}}

\newcommand{\rLambda}{\mathrm{\Lambda}}
\newcommand{\rSigma}{\mathrm{\Sigma}}

\newcommand{\vLambda}{\bm{\rLambda}}
\newcommand{\vSigma}{\bm{\rSigma}}

\makeatletter
\newcommand*\bdot{\mathpalette\bdot@{.7}}
\newcommand*\bdot@[2]{\mathbin{\vcenter{\hbox{\scalebox{#2}{$\m@th#1\bullet$}}}}}
\makeatother

\makeatletter
\DeclareRobustCommand\onedot{\futurelet\@let@token\@onedot}
\def\@onedot{\ifx\@let@token.\else.\null\fi\xspace}

\def\eg{\emph{e.g}\onedot} \def\Eg{\emph{E.g}\onedot}
\def\ie{\emph{i.e}\onedot} \def\Ie{\emph{I.e}\onedot}
\def\cf{\emph{cf}\onedot} \def\Cf{\emph{Cf}\onedot}
\def\etc{\emph{etc}\onedot} \def\vs{\emph{vs}\onedot}
\def\wrt{w.r.t\onedot} \def\dof{d.o.f\onedot} \def\aka{a.k.a\onedot}
\def\etal{\emph{et al}\onedot}
\makeatother

\abstract
{This paper investigates the impact of various data augmentation techniques on the performance of object detection models. Specifically, we explore classical augmentation methods, image compositing, and advanced generative models such as Stable Diffusion XL and ControlNet. The objective of this work is to enhance model robustness and improve detection accuracy, particularly when working with limited annotated data. Using YOLOv8, we fine-tune the model on a custom dataset consisting of commercial and military aircraft, applying different augmentation strategies. Our experiments show that image compositing offers the highest improvement in detection performance, as measured by precision, recall, and mean Average Precision (mAP@0.50). Other methods, including Stable Diffusion XL and ControlNet, also demonstrate significant gains, highlighting the potential of advanced data augmentation techniques for object detection tasks. The results underline the importance of dataset diversity and augmentation in achieving better generalization and performance in real-world applications. Future work will explore the integration of semi-supervised learning methods and further optimizations to enhance model performance across larger and more complex datasets.}

\onecolumn \maketitle \normalsize \setcounter{footnote}{0} \vfill

\section{\uppercase{Introduction}}
\label{sec:introduction}
Deep learning models, particularly Convolutional Neural Networks (CNNs) have revolutionized the field of computer vision, achieving state-of-the-art performance on a wide range of tasks, including image classification and object detection. However, the performance of these models is heavily reliant on the availability of large, high-quality datasets. In many real-world scenarios, obtaining sufficient training data can be challenging, especially for specific domains or rare classes.

To address this limitation, data augmentation has been shown to produce promising ways to increase the accuracy of classification tasks, to artificially expand training datasets. Previous research has explored various data augmentation techniques such as traditional methods, such as rotation, flipping, and cropping \cite{perez2017effectivenessdataaugmentationimage}, and generative adversarial networks (GANs) to generate synthetic images \cite{8388338}. Some other works have changed images' semantics using an off-the-shelf diffusion model, which generalizes to novel visual concepts from a few labeled examples \cite{trabucco2023effectivedataaugmentationdiffusion}, another study has used Multi-stage Augmented Mixup (MiAMix), which integrates image augmentation into the mixup framework, utilizes multiple diversified mixing methods concurrently, and improves the mixing method by randomly selecting mixing mask augmentation methods \cite{pr11123284}.

One specific area that faces the challenge of scarce labeled data is aircraft detection. Accurate and timely identification of aircraft is crucial in various sectors, including airspace security, airport traffic management, and military applications \cite{Arwin:09}. 

In this paper, we propose a novel data augmentation method for this application that combines elements from multiple images to create a new, synthetic image, which we will refer to as Image Compositing. Impressively, we show that our method outperforms other complex generative model techniques such as multi-modal diffusion models \cite{sd}. 
\section{\uppercase{Related Work}}
\label{sec:methodology}
Data augmentation and synthetic data generation have emerged as powerful techniques to enhance the performance and robustness of deep learning models, particularly in scenarios with limited data.

\subsection{Data augmentation methods}
Data augmentation techniques have been widely employed to enhance the performance and generalization of deep learning models, especially in scenarios with limited data. Traditional methods, such as geometric transformations (e.g., random cropping, flipping, rotation) and colour jittering, have been effective in improving model robustness \cite{traditional}.

Recent advancements in data augmentation have focused on more sophisticated techniques. For instance, RICAP \cite{random_cropping} randomly crops and patches images to create new training examples, while also mixing class labels to introduce soft label learning. This approach has shown promising results in various computer vision tasks.

To address the issue of colour variations between different cameras, a novel approach has been proposed to map colour values using deep learning \cite{colour}. By learning colour-mapping parameters, this technique enables the augmentation of colour data by converting images from one camera to another, effectively expanding the training dataset.

Another recent technique, SmoothMix, addresses the limitations of existing regional dropout-based data augmentation methods \cite{smooth_mix}. By blending images based on soft edges and computing corresponding labels, SmoothMix minimizes the "strong-edge" problem and improves model performance and robustness against image corruption.

In the domain of hyperspectral image (HSI) denoising, data augmentation has been less explored. A new method called PatchMask has been proposed to augment HSI data while preserving spatial and spectral information \cite{patch_mask}. By creating diverse training samples that lie between clear and noisy images, PatchMask can enhance the effectiveness of HSI denoising models.

Recent advancements in attention mechanisms have enabled more effective data augmentation techniques. Attentive CutMix \cite{cut_mix} is a novel method that leverages attention maps to identify the most discriminative regions within an image, and then selectively applies cut-mix operations to these regions. This targeted approach can lead to significant improvements in model performance.

\subsection{Synthetic Data generation methods}
Synthetic data generation has emerged as a powerful technique to address data scarcity and domain shift challenges in various domains. By generating realistic synthetic data, models can be trained on larger and more diverse datasets, leading to improved performance.

Generative Adverserial Networks (GANs) have gained widespread popularity for their ability to produce high-quality synthetic data by training a generator to create realistic samples while a discriminator distinguishes between real and generated data. Their versatility has been demonstrated across domains such as image synthesis \cite{sodgan} and industrial object detection \cite{cyclegan}. However, GANs can be challenging to train and often suffer from mode collapse, where the generator fails to capture the full diversity of the data distribution. 

Variational Autoencoders (VAEs) learn a latent representation of the data distribution and can generate new data points by sampling from this latent space. VAEs are more stable to train than GANs, but they often produce lower-quality samples, especially for complex data distributions. VAEs have been applied to various tasks, including image generation, anomaly detection, and data augmentation. For example, VAEs have been used to generate synthetic medical images for training medical image segmentation models \cite{vae-gan} and to synthesize semantically rich images for geospatial applications \cite{vae-info-cgan}.

Vector Quantised-Variational Autoencoders (VQ-VAEs) enhance the capabilities of VAEs by introducing a discrete latent code, making it more efficient and interpretable. VQ-VAEs have been shown to be effective in generating high-quality images and can be used as a building block for more complex generative models. VQ-VAEs have been applied to various tasks, including image compression, image generation, and video prediction. For example, VQ-VAEs have been used to generate synthetic data for human activity recognition (HAR) with complex multi-sensor inputs \cite{vq-vae}.

Diffusion models gradually denoise a random noise vector to generate realistic data samples. Recent work, such as \cite{ho2020denoising}, has shown that diffusion models can achieve state-of-the-art results in image generation. Diffusion models have been applied to various tasks, including image generation, image restoration, and text-to-image generation. Latent diffusion models (LDMs) \cite{synthesis-diffusion} enhance efficiency by operating in a compressed latent space, significantly reducing computational costs.
\section{\uppercase{Background}}
\label{sec:background}

\subsection{Data Collection}
\label{sub:data-collection}
While existing datasets like FGVC-Aircraft provide valuable resources for aircraft recognition, they primarily focus on aircraft images captured from aerial perspectives, which do not align with the specific requirements of ground-based aircraft detection. To address this limitation, we adopted a novel data curation strategy involving a multi-step process.

We meticulously sourced images from various online platforms, including stock photo websites and aviation enthusiast forums such as \href{jetphotos.com}{JetPhotos}. This approach allowed us to gather a diverse collection of images capturing aircraft in various scenarios, with a specific focus on ground-based perspectives.

To efficiently label the large dataset, we employed a semi-automated approach leveraging the Grounding DINO model using Roboflow. This model was trained on a large-scale image-text dataset and can accurately localize objects in images given textual prompts. By providing a prompt such as "plane", the model was able to generate initial bounding box proposals.

However, to ensure high-quality annotations, each image with proposed bounding boxes was then carefully examined. Incorrect or missed detections were corrected, and additional annotations were added as needed. The final dataset split can be seen in \autoref{tab:baseline-split}.

\begin{table}
\small
\centering
\caption{Baseline Dataset Split}

\setlength\tabcolsep{4pt}
\begin{tabular}{lccc}
\toprule
Class & Training&Validation&Test\\ \midrule
Commercial &218 & 62 & 36    \\
Military& 22 & 9 & 6 \\ 

\bottomrule
\end{tabular}
\vspace{0pt}

\label{tab:baseline-split}
\end{table}

The following sections will first explore the baseline augmentation techniques -- classical data augmention Stable Diffusion and its extension, ControlNet. Building upon these foundations, we will then introduce a novel method for data augmentation: Image Compositing.
\subsection{Baselines Methods}
\paragraph {Classical Data Augmentation} These methods used were horizontal flipping, Gaussian blurring and exposure adjustment. Horizontal flipping was applied to introduce spatial variability. This technique mirrors the image along the vertical axis, effectively doubling the dataset size without altering the underlying semantic content. Gaussian blurring introduces a controlled level of noise and blurring, mimicking the effects of atmospheric conditions or sensor noise. Additionally, exposure adjustment was employed to vary the overall intensity of the image, simulating changes in illumination.

\paragraph{Stable Diffusion XL}\label{subsec:sdxl}Stable Diffusion XL is a state-of-the-art text-to-image model capable of generating highly realistic and detailed images from textual descriptions \cite{sdxl}. We provided specific prompts, such as "A photo of a military plane in sky, taken from the ground" or "A photo of a commercial plane in sky, taken from the ground," as well as negative prompts such as "cropped, close-up, low resolution, blurred, partial view, cut-off edges," to ensure they met our specific requirements. The generated images were then labeled using the approach in section \ref{sub:data-collection}.

\paragraph{Stable Diffusion XL with ControlNet}We had provided the Stable Diffusion XL model with a guidance image to influence its output, ensuring that the generated images were consistent with the desired characteristics. This was carried out using the recently published model of ControlNet \cite{controlnet}. The idea of ControlNet is to to use a conditioning input such as a segmentation maps, Canny edges and depth maps that can be used to control the generated image. In our work we utilized Canny edges as the conditioning input for the ControlNet. We used a subset of the training images and obtained their Canny edge images by applying Canny edge detection. Then we feed these Canny edge images as input to the network and in a similar way to \autoref{subsec:sdxl} we provided a prompt that will generate a plane. Our hypothesis is that using the Canny edges and the ControlNet will force the Stable Diffusion XL model to generate plane exactly in the same location as the original input images. In this way we will be able to use the original bounding box information to fine-tune our plane detector. 

\subsection{Image Compositing}
\begin{figure*}[h]
\begin{center}
   \includegraphics[width=\textwidth, height = 0.22\textwidth]{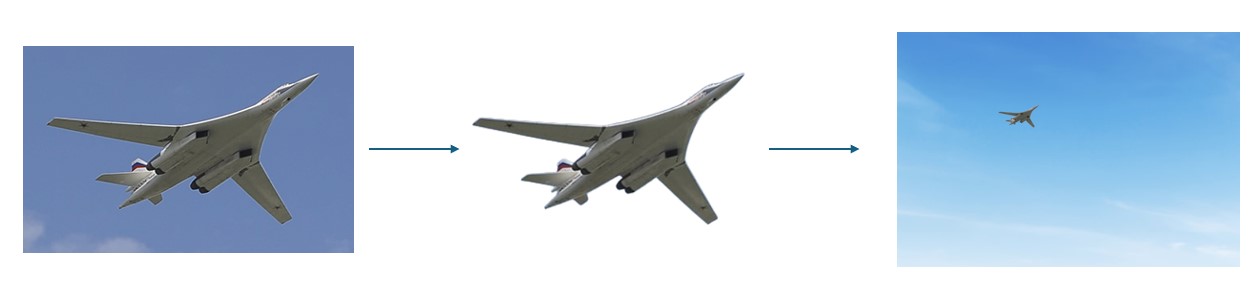}
\end{center}
   \caption{Data Generation Using Image Composition}
\label{fig:6}
\end{figure*}

Image fusion techniques were employed, which involved background removal, sky integration and seam reduction, illustrated in \autoref{fig:6}. Background elements were firstly removed from images containing an aircraft, isolating the foreground object --- the aircraft. The foreground aircraft objects were then integrated onto sky background images captured from a ground perspective. The foreground aircraft was then rotated by an angle between $0^\circ$ to $10^\circ$ and flipped horizontally, increasing robustness of training data. To enhance image realism, Gaussian filtering was applied to blur the boundaries between the foreground aircraft and the background sky, minimising visible seams.

\paragraph{Gaussian Filtering} An image processing technique employed for noise reduction and image smoothing. This is accomplished by applying a filter kernel whose weights are defined by a Gaussian function. This function is a bell-shaped curve that assigns higher weights to pixels closer to the center and progressively lower weights to those further away. 

The Gaussian filter is applied to an image by convolving the Gaussian kernel with the image. 
\section{\uppercase{Methodology}}
\label{sec:methodology}
\subsection{Problem definition}
We define a labeled baseline image dataset $ D_{\base} =\ (\vx_i,\vy_i)$, where $\vx_i $ represents the $i^{th}$  image and $\vy_i$ represents the corresponding class label of image $\vx_i $. This dataset comprises images of commercial and military planes. The dataset is partitioned into three splits: the training set split, ${D_{\train}=\ ({\vx}_i,\vy_i})$, the validation set split,  $ {D_{\val}=\ ({\vx}_i,\vy_i})$ and testing set split, ${D_{\test}=\ ({\vx}_i,\vy_i}).$ We use $D_{\train}$ to train a neural network, that consists of a backbone $f_{\theta}$ and a classifier $g_{\phi}$ (last layer of the network). 

The validation set $D_{\val}$ is used in order to save the model with the highest validation accuracy. Finally, we use $D_{\test}\ $ to calculate the test set classification accuracy. 

\subsection{Training Phase}
In each batch, we use training images along with corresponding annotations. Let $D = \{\mathbf{x}_i, \mathbf{y}_i\}_{i=1}^N$ denote the dataset, where $\mathbf{x}_i$ represents the input image, and $\mathbf{y}_i$ represents the corresponding annotations.

During the forward pass, the model predicts $\hat{y}_i$ for each input $\mathbf{x}_i$. The prediction $\hat{y}_i$ includes the bounding box coordinates, class probabilities, and objectness score. 

The total loss, $\mathcal{L}_\text{total}$, is calculated for each batch as:
\begin{equation}
    \mathcal{L}_\text{total} = \mathcal{L}_\text{obj} + \mathcal{L}_\text{cls} + \mathcal{L}_\text{bbox},
\end{equation}
where:
\begin{itemize}
    \item $\mathcal{L}_\text{obj}$ is the objectness loss.
    \item $\mathcal{L}_\text{cls}$ is the classification loss.
    \item $\mathcal{L}_\text{bbox}$ is the bounding box regression loss.
\end{itemize}

The weights of the network, $\mathbf{W}$, are updated using the Adam optimizer:
\begin{equation}
    \mathbf{W}_{t+1} = \mathbf{W}_t - \eta \nabla \mathcal{L}_\text{total},
\end{equation}
where $\eta$ is the learning rate.

Early stopping is employed to halt training if the validation loss does not improve for $p$ consecutive epochs (patience $p = 10$). The training process is summarized in Algorithm \ref{alg:train}.

\begin{algorithm}[!h]
 \caption{YOLOv8 Training Process}
 \label{alg:train}
 \KwData{Dataset $D$, configuration file $data.yaml$, pre-trained model $yolov8s.pt$, number of epochs $E=500$, patience $P=10$}
 \KwResult{Trained model with updated weights}
 
 Initialize model with pre-trained weights $yolov8s.pt$\;
 Set training parameters: $batch\_size = 16$, $epochs = 500$, $learning\_rate = 0.001667$, $optimizer = \text{AdamW}$\;
 Set data configuration file path: $data.yaml$\;
 
 \For{$epoch \gets 1$ \textbf{to} $E$}{
  \For{each batch $B$ in the training set $D$}{
   Perform forward pass on batch $B$\;
   Calculate loss using classification, localization, and confidence components\;
   Perform backward pass and update model weights using AdamW optimizer\;
  }
  
  Calculate validation loss on validation set\;
  \If{validation loss does not improve for $P$ epochs}{
   Save the model with the lowest validation loss\;
   \textbf{Break}\;
  }
 }
 \textbf{Return} the trained model with optimized weights\;
\end{algorithm}

\subsection{Inference Stage}
During the inference stage, the model processes each test image $\mathbf{x}_i$ from the test dataset $D_\text{test} = \{\mathbf{x}_j\}_{j=1}^M$ to predict the class labels and bounding boxes. The predicted class label $\hat{\mathbf{y}}_i$ for each detected object is derived as:
\begin{equation}
    \hat{\mathbf{y}}_i = \arg\max_{k \in [C]} \hat{p}_{ik},
\end{equation}
where $\hat{p}_{ik}$ is the predicted probability for class $k$, and $C$ is the total number of classes. 

The bounding box predictions are represented as $\hat{\mathbf{b}}_i = (\hat{x}_i, \hat{y}_i, \hat{w}_i, \hat{h}_i)$, where $\hat{x}_i$ and $\hat{y}_i$ denote the center coordinates of the bounding box, and $\hat{w}_i$ and $\hat{h}_i$ are its width and height. The model leverages anchor-free mechanisms to predict these bounding boxes directly at specific feature map locations, reducing the reliance on predefined anchor boxes. The bounding boxes are computed through the regression head of the network, which predicts the normalized offsets for each feature map grid cell corresponding to the detected objects.

To refine the predictions, the model applies post-processing techniques such as non-maximum suppression (NMS) to eliminate redundant bounding boxes and retain only the most confident detections. This is mathematically expressed as:
\begin{equation}
    \hat{\mathbf{b}}_i = \text{NMS}(\{\mathbf{b}_{ij}\}_{j=1}^N, \{\hat{p}_{ij}\}_{j=1}^N, \tau),
\end{equation}
where $\{\mathbf{b}_{ij}\}_{j=1}^N$ and $\{\hat{p}_{ij}\}_{j=1}^N$ are the sets of predicted bounding boxes and their associated confidence scores for image $\mathbf{x}_i$, and $\tau$ is the IoU threshold used to filter overlapping boxes.

To evaluate the model's performance, we utilize three key metrics:
\paragraph{Mean Average Precision at IoU 0.50 (mAP@0.50)} This metric evaluates the overall detection performance by calculating the average precision across all classes for a fixed Intersection-over-Union (IoU) threshold of 0.50.
\paragraph{Precision} Defined as the ratio of true positive detections to the sum of true positives and false positives. It measures how many of the predicted detections are relevant.
\paragraph{Recall} Defined as the ratio of true positive detections to the total number of ground-truth instances. It measures the model's ability to detect relevant objects.

These metrics collectively provide a comprehensive evaluation of the model's performance, capturing its precision, completeness, and overall detection capability.

\section{\uppercase{experiments}}
\label{sec:experiments}

\subsection{Setup}
\paragraph{Datasets}The final augmented dataset can be seen in \autoref{tab:augmented-split}.

\begin{figure*}[t]
    \centering
        \caption*{Generated Images from Stable Diffusion XL}
    \begin{minipage}{\textwidth}
        \centering
        \begin{minipage}{0.18\textwidth}
            \centering
            \includegraphics[width=\textwidth]{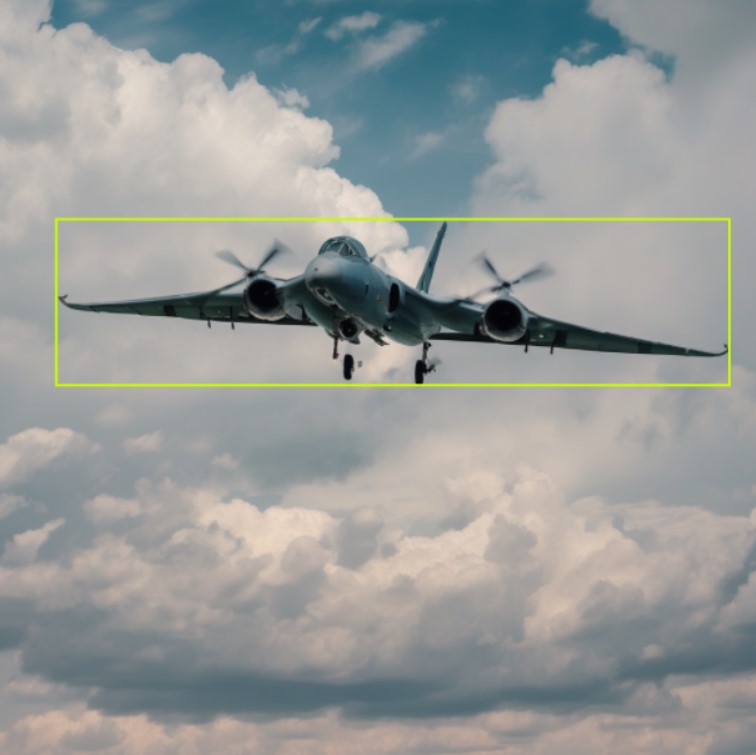}
        \end{minipage} \hfill
        \begin{minipage}{0.18\textwidth}
            \centering
            \includegraphics[width=\textwidth]{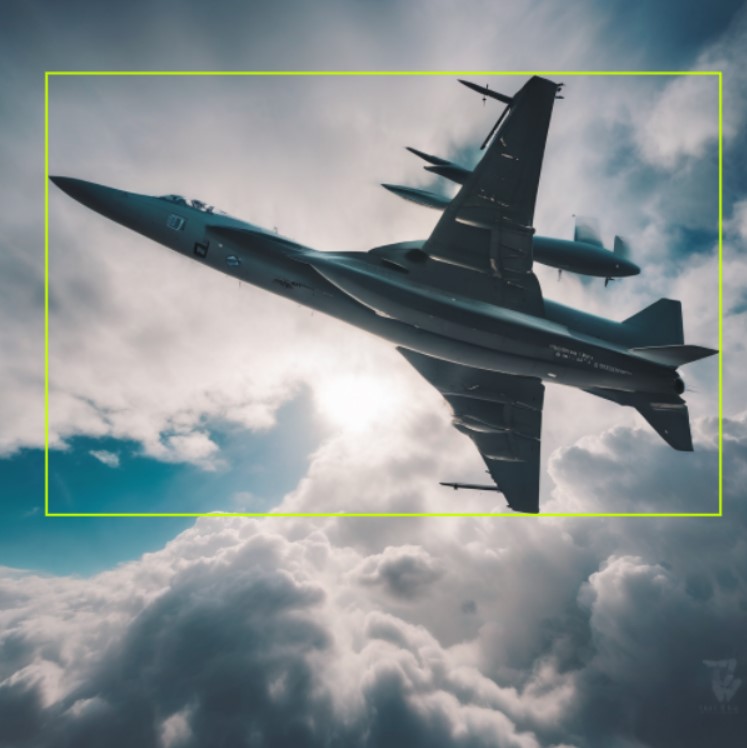}
        \end{minipage} \hfill
        \begin{minipage}{0.18\textwidth}
            \centering
            \includegraphics[width=\textwidth]{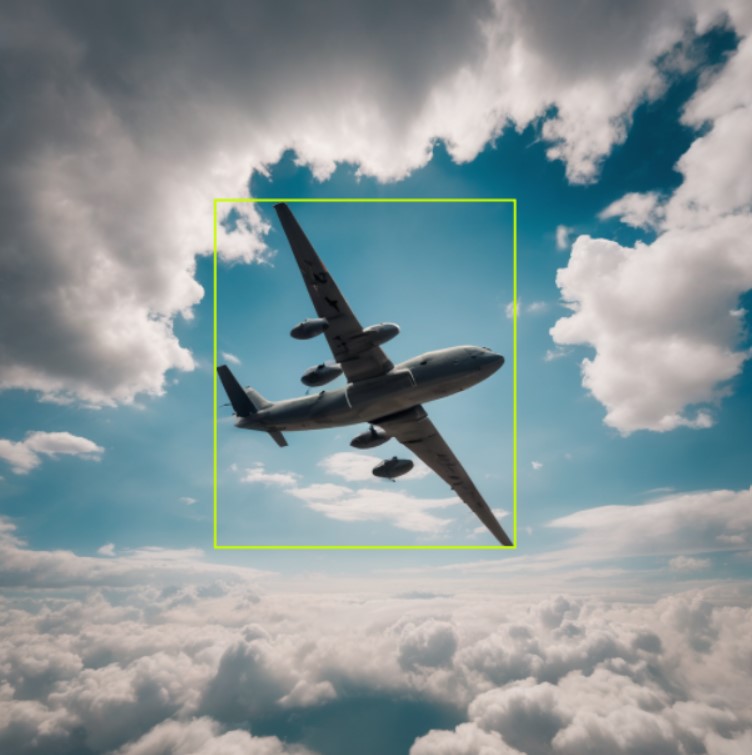}
        \end{minipage} \hfill
        \begin{minipage}{0.18\textwidth}
            \centering
            \includegraphics[width=\textwidth]{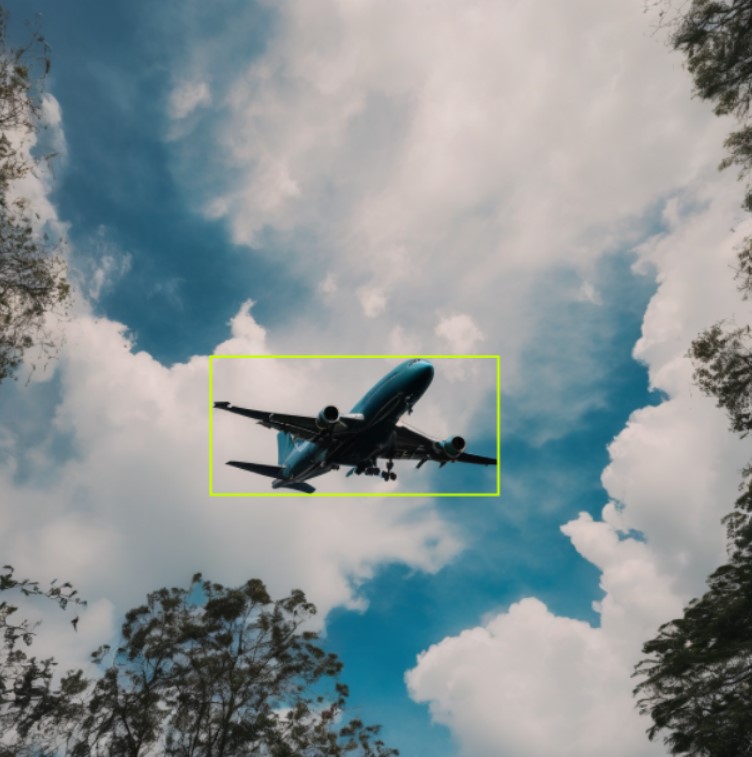}
        \end{minipage} \hfill
        \begin{minipage}{0.18\textwidth}
            \centering
            \includegraphics[width=\textwidth]{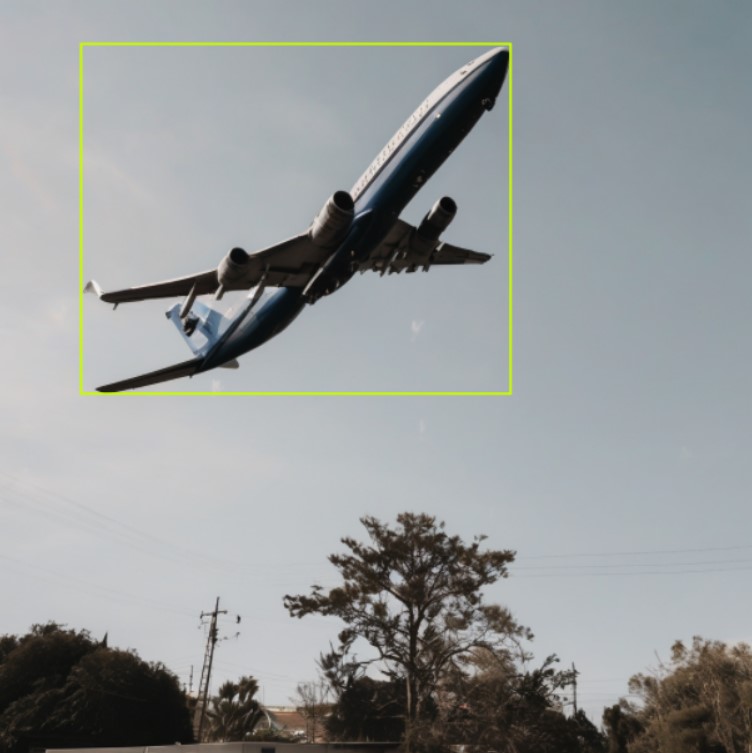}
        \end{minipage}
    \end{minipage}

    \vspace{1em} 
    \caption*{Generated Images from Stable Diffusion + ControlNet}
    \begin{minipage}{\textwidth}
        \centering
        \begin{minipage}{0.18\textwidth}
            \centering
            \includegraphics[width=\textwidth]{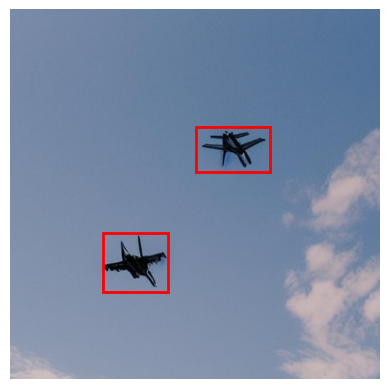}
        \end{minipage} \hfill
        \begin{minipage}{0.18\textwidth}
            \centering
            \includegraphics[width=\textwidth]{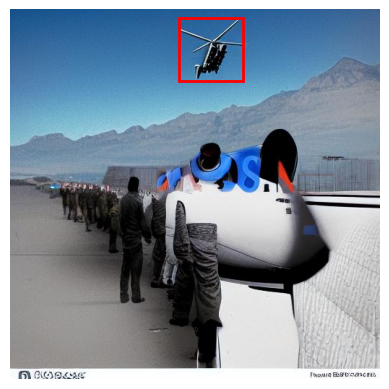}
        \end{minipage} \hfill
        \begin{minipage}{0.18\textwidth}
            \centering
            \includegraphics[width=\textwidth]{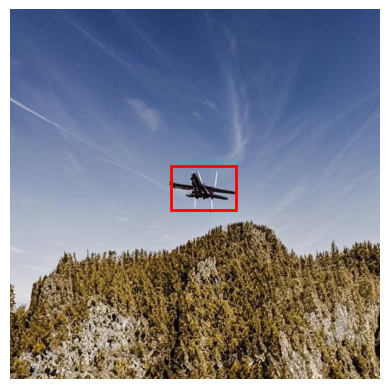}
        \end{minipage} \hfill
        \begin{minipage}{0.18\textwidth}
            \centering
            \includegraphics[width=\textwidth]{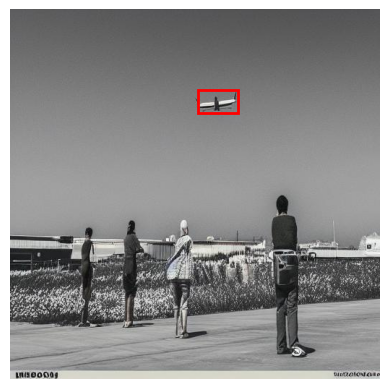}
        \end{minipage} \hfill
        \begin{minipage}{0.18\textwidth}
            \centering
            \includegraphics[width=\textwidth]{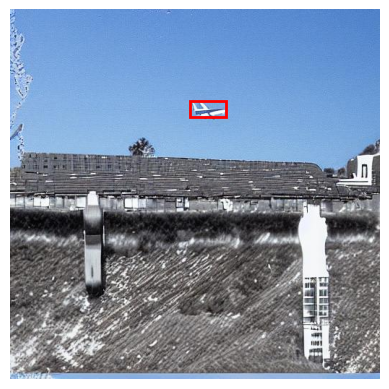}
        \end{minipage}
    \end{minipage}
    
    \vspace{1em} 

    \caption*{Generated Images from Image Composition}
    \begin{minipage}{\textwidth}
        \centering
        \begin{minipage}{0.18\textwidth}
            \centering
            \includegraphics[width=\textwidth]{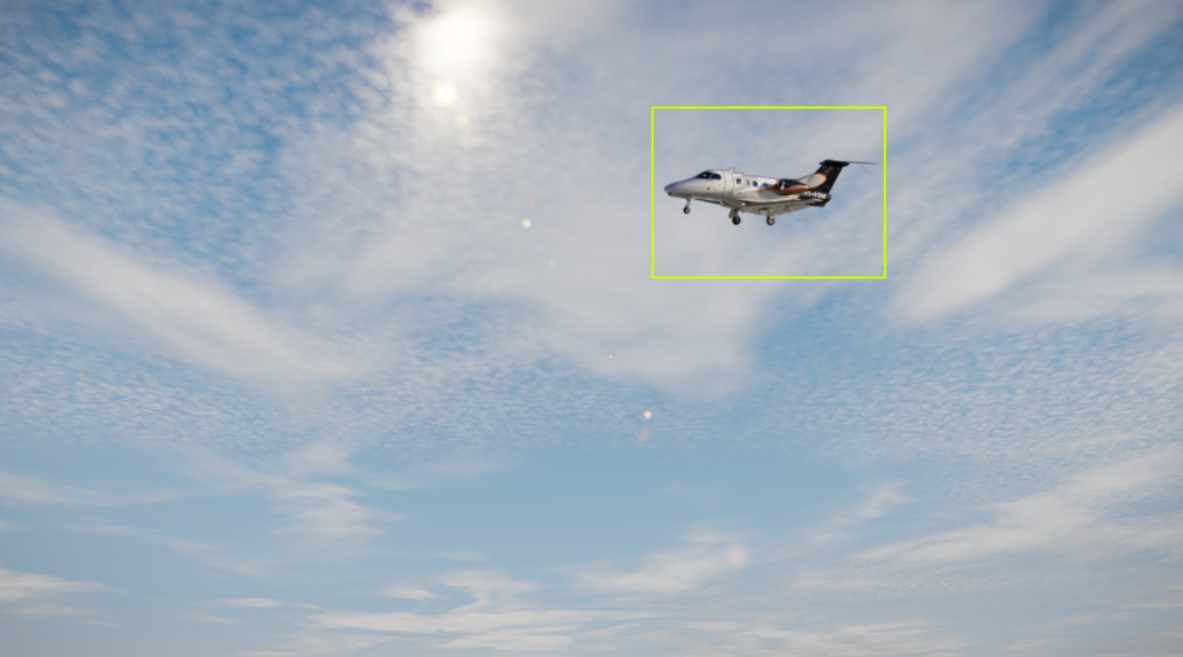}
        \end{minipage} \hfill
        \begin{minipage}{0.18\textwidth}
            \centering
            \includegraphics[width=\textwidth]{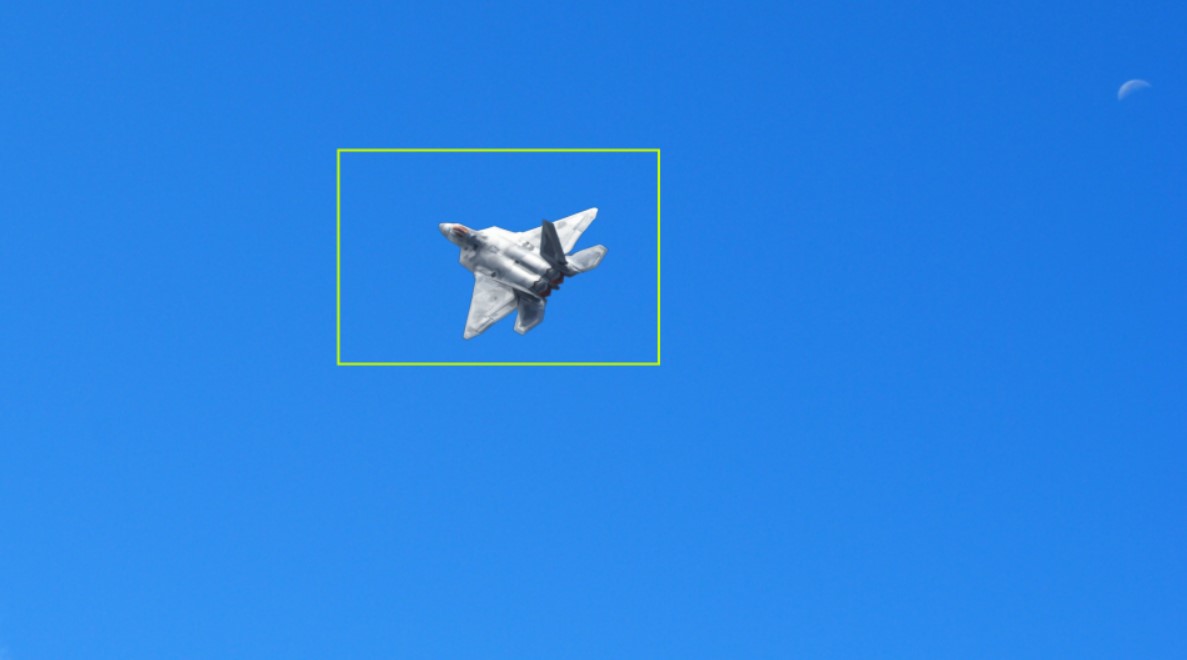}
        \end{minipage} \hfill
        \begin{minipage}{0.18\textwidth}
            \centering
            \includegraphics[width=\textwidth]{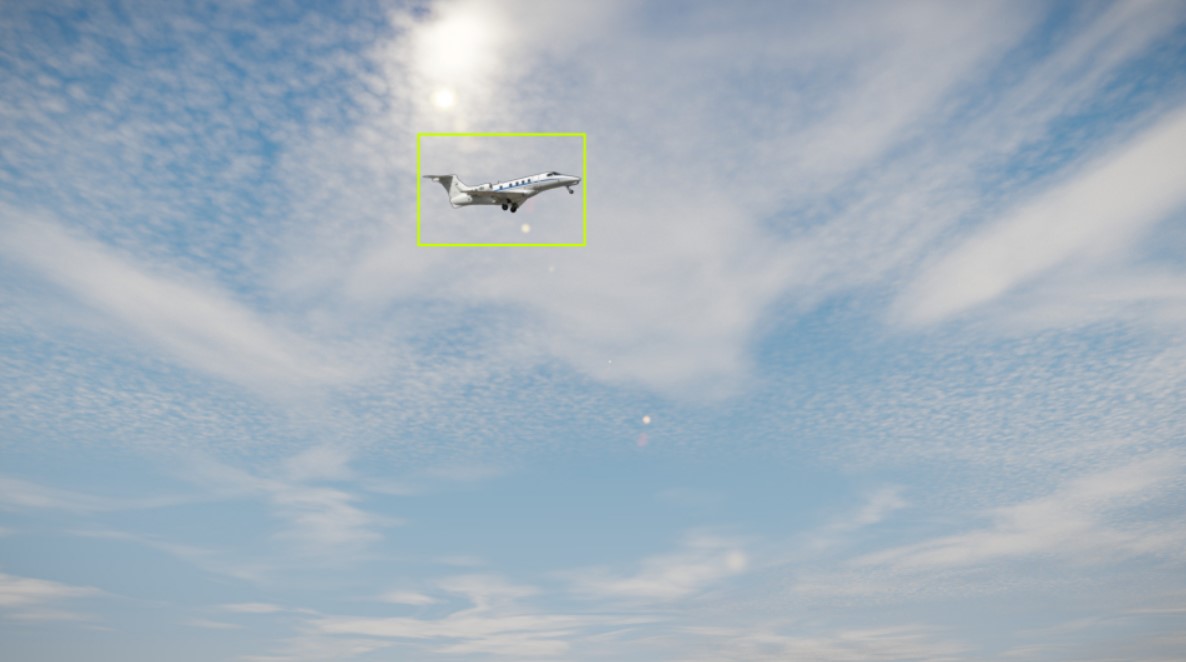}
        \end{minipage} \hfill
        \begin{minipage}{0.18\textwidth}
            \centering
            \includegraphics[width=\textwidth]{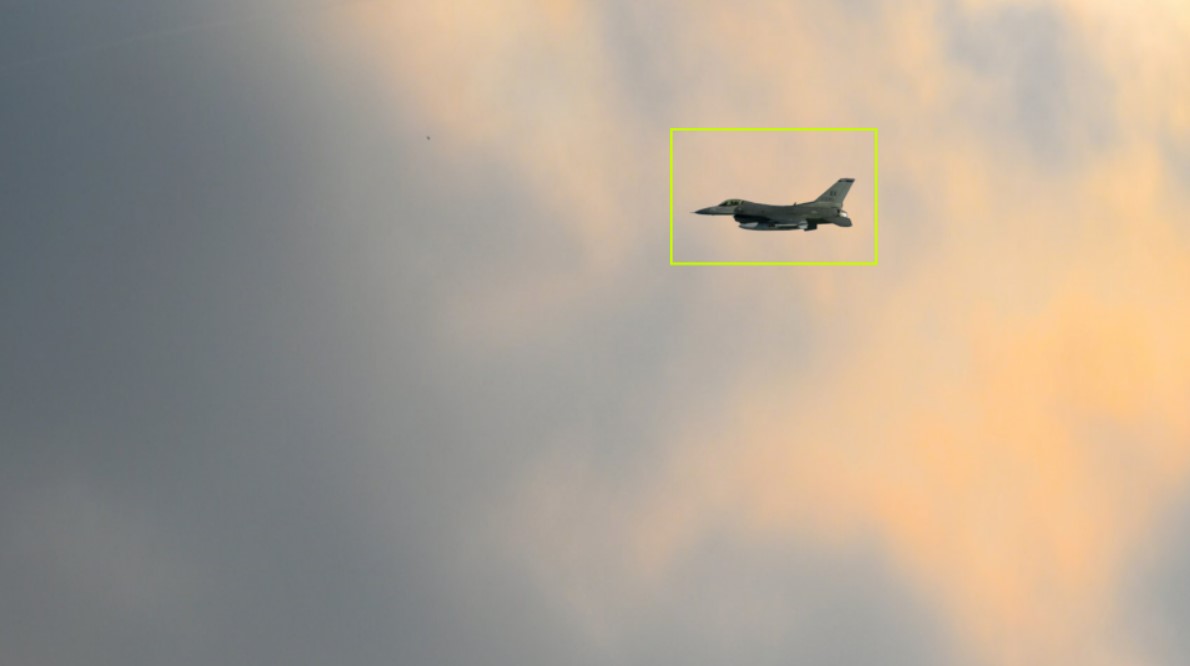}
        \end{minipage} \hfill
        \begin{minipage}{0.18\textwidth}
            \centering
            \includegraphics[width=\textwidth]{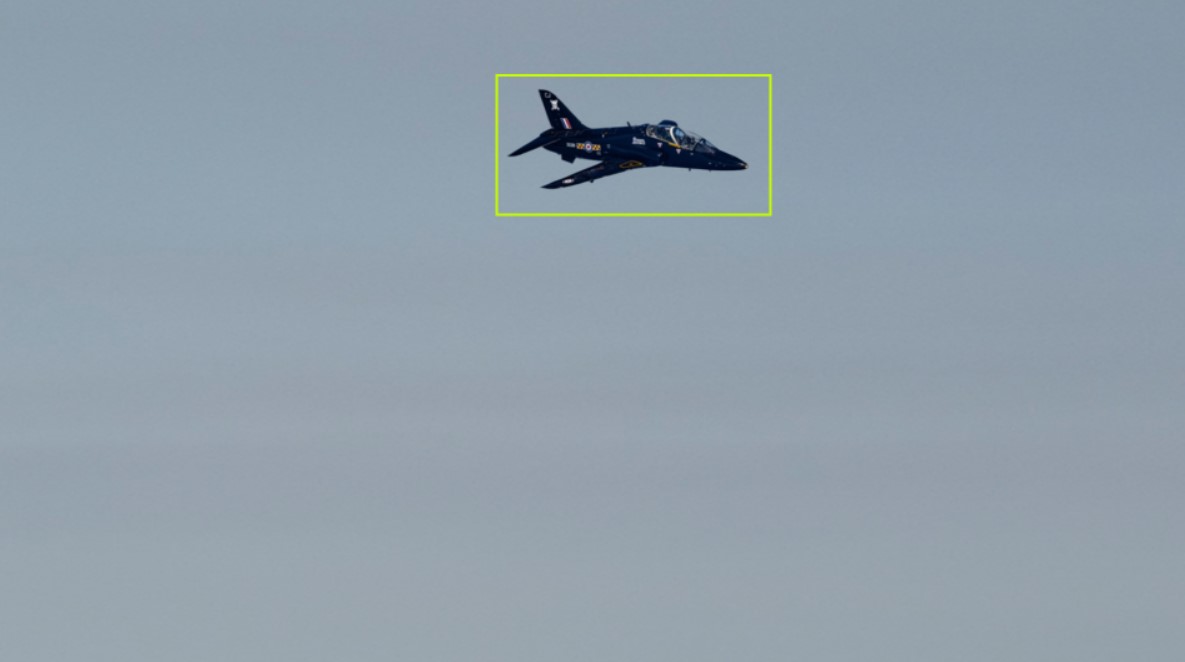}
        \end{minipage}
    \end{minipage}
    \vspace{1em}
    \caption{Images Generated using Different Methods}
    \label{fig:generated_images}
\end{figure*}

\begin{table}
\small
\centering
\caption{Augmented Dataset Split}

\setlength\tabcolsep{4pt}
\begin{tabular}{lccc}
\toprule
Class & Training&Validation&Test\\ \midrule
Commercial &307 & 73 & 36    \\
Military& 338 & 43 & 6 \\ 
\bottomrule
\end{tabular}
\vspace{0pt}

\label{tab:augmented-split}
\end{table}

\paragraph{Network} In our work, we had used YOLOv8 and had fine-tuned for our custom datasets. YOLOv8 employs pretrained backbones such as CSPDarknet53. These pretrained weights, $\mathbf{W}_\text{pretrained}$, initialize the model to improve convergence.

\paragraph{Implementation Details}
Our implementation is based on Python. We utilized the Ultralytics YOLOv8 framework for model training and inference, with PyTorch serving as the underlying deep learning library for GPU-accelerated computations. For image augmentation and preprocessing tasks, OpenCV, NumPy, \texttt{diffusers} and \texttt{transformers} libraries.

\paragraph{Hyperparameters}
In the experiment, the batch size is implicitly set by the available GPU memory, but it typically defaults to 16 for optimal performance. The model was trained for 500 epochs, with early stopping enabled by setting the patience to 10. The AdamW optimizer was used, with a learning rate of 0.001667 and momentum of 0.9. The optimizer was configured with parameter groups, where different decay rates were applied to various parts of the model. Specifically, the weight decay for the first group of parameters was set to 0.0 (no weight decay), while for the second group (weights), the decay was set to 0.0005. The third group (bias parameters) had a weight decay of 0.0, ensuring that bias terms did not undergo regularization. The loss function employed is a combination of objectness, classification, and bounding box regression losses, tailored for object detection tasks. We use the validation set to calculate the validation accuracy and save the model with the highest validation accuracy through comparisons at each epoch. The performance of the model is monitored using metrics such as mean Average Precision (mAP), precision, and recall, which are calculated at each epoch to track the model’s detection accuracy on the test set.

\subsection{Ablation Study}
In this study, we conducted a series of experiments to evaluate the impact of different data augmentation techniques on model performance. For each dataset, we applied a distinct augmentation method and compared it to the baseline, which involved training the model on the original unmodified data without any augmentation. The performance of each approach was assessed using mAP@0.50, precision, and recall. These metrics were calculated for each augmented dataset and presented in a comparative manner in \autoref{tab:augmentation-results}.

\begin{figure*}[ht]
  \centering
\includegraphics[width=0.9\textwidth]{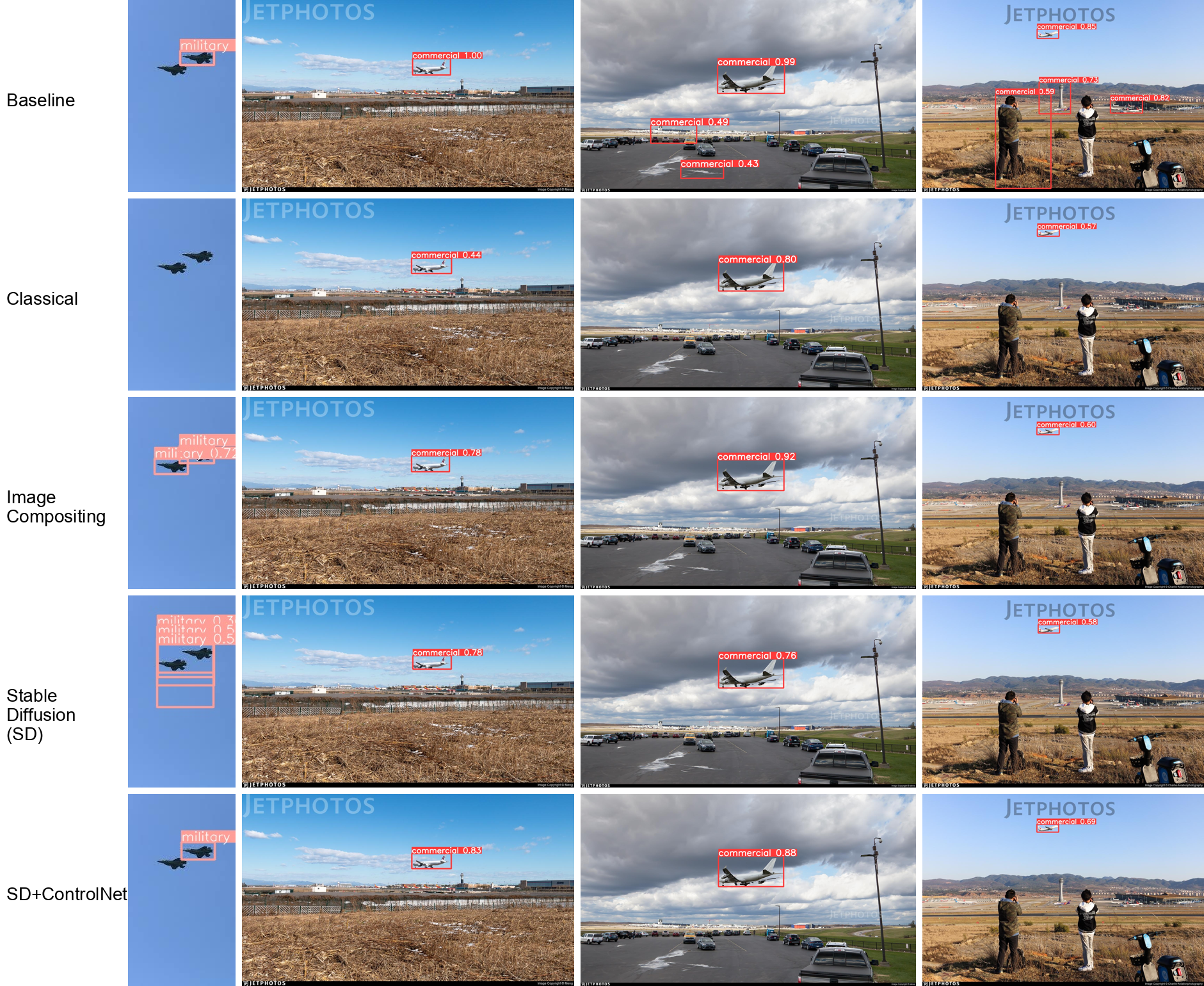}  
\caption{Prediction Results for Each Augmented Dataset}
  \label{fig:res}
 \end{figure*}
\subsection{Performance Comparison}
\autoref{tab:augmentation-results} provides a quantitative comparison of precision, recall, and mAP@0.50 metrics across different data augmentation techniques. The baseline model, trained on the original dataset without augmentation, showed moderate performance with an mAP@0.50 of 0.654. In contrast, classical data augmentation techniques such as flipping and blurring significantly improved performance, achieving an mAP@0.821. The proposed Image Compositing method outperformed all other techniques, with the highest mAP@0.911, precision of 0.904, and recall of 0.907. \autoref{fig:generated_images} visually supports these findings by showcasing sample images generated through each augmentation method. The superior performance of Image Compositing when compared to advanced generative models like Stable Diffusion can be attributed to the distribution shift between the source images and the images generated by Stable Diffusion. This shift is evident in \autoref{fig:generated_images}, particularly in the first row, where the airplanes in the images generated by Stable Diffusion noticeably differ from the airplanes in our dataset. 

\autoref{fig:res} visually corroborates the quantitative results presented in \autoref{tab:augmentation-results}. The baseline model exhibits a high number of missed detections and incorrect labeling, resulting in a low precision score as shown in \autoref{tab:augmentation-results}. The classical augmentation method showed an improvement over the baseline, with a notable increase in detection accuracy. However, some aircraft remain undetected, aligning with the higher recall score compared to the original dataset. 

Image compositing gives the best results with accurate and confident bounding box predictions for all aircraft. The model effectively handles cluttered backgrounds and distant objects, which is consistent with the scores in \autoref{tab:augmentation-results}. While showcasing improved performance over the original dataset, the model trained with Stable Diffusion showed some inconsistencies in the bounding box predictions, aligning with its scores which are higher than the baseline but lower than Image Compositing. Stable Diffusion + ControlNet has a balance between precision and recall, but still falls slightly short of the performance achieved by Image Compositing, as evidenced by the scores in \autoref{tab:augmentation-results}.

\begin{table}
\small
\centering
\caption{Performance Comparison of Different Augmentation Methods}

\setlength\tabcolsep{2pt}
\begin{tabular}{lcccc}
\toprule
Method & Precision&Recall&mAP50& Epoch\\ \midrule
Original &0.558 & 0.699 &  0.654 & 2    \\
Classical& 0.856 & 0.794 &0.821& 28 \\ 
Image Compositing &0.904 & 0.907 & 0.911 & 32 \\ 
Stable Diffusion (SD) &0.718 & 0.809 &0.808& 25   \\ 
SD+ControlNet  & 0.874 & 0.703 &0.854 & 37  \\ 

\bottomrule
\end{tabular}
\vspace{0pt}

\label{tab:augmentation-results}
\end{table}

\subsection{Verification of Hypotheses}
The experiments were designed to validate that advanced augmentation methods, including generative models, would improve object detection performance over classical methods and that Image Compositing, as a novel augmentation strategy, would outperform state-of-the-art generative models in both precision and recall.

The results supported both hypotheses. Stable Diffusion XL and Stable Diffusion XL with ControlNet demonstrated significant performance gains (mAP@0.808 and mAP@0.854, respectively) over the baseline model, confirming the effectiveness of advanced augmentation methods. 
Moreover, the superior performance of Image Compositing across all metrics validated its position as the most effective augmentation method tested.

\section{\uppercase{conclusion}}
\label{sec:conclusion}
In this research, we proposed a comprehensive framework for improving object detection performance using various data augmentation techniques. Our approach leverages a combination of classical augmentation methods, image compositing, and advanced models like Stable Diffusion XL and ControlNet to augment the dataset. By augmenting the dataset in different ways, we were able to improve model robustness and generalization, addressing the challenges of limited annotated data in object detection tasks. 

Through rigorous experiments on a custom dataset involving both commercial and military aircraft, we demonstrated that different augmentation techniques provide varying degrees of improvement in detection accuracy, as measured by precision, recall, and mAP@0.50. Among the methods evaluated, image compositing stood out as the most effective in terms of performance, achieving the highest precision and recall scores, as well as the best mAP. 

Our results validate the hypothesis that data augmentation can significantly enhance the performance of object detection models, even in the presence of complex and imbalanced datasets. Moving forward, we plan to further refine and optimize the augmentation strategies, combining them with cutting-edge techniques such as generative adversarial networks and semi-supervised learning methods. Additionally, extending our approach to larger datasets and applying it across other domains, such as autonomous vehicles and medical imaging, presents an exciting direction for future work. Our ultimate goal is to continue advancing the state-of-the-art in object detection, improving both model accuracy and computational efficiency.

\bibliographystyle{apalike}
{\small
\bibliography{example}}

\end{document}